\documentclass[letterpaper, 10 pt, journal, twoside]{IEEEtran}
\usepackage{amsmath,amsfonts}
\usepackage{algorithmic}
\usepackage{algorithm}
\usepackage{array}
\usepackage[caption=false,font=normalsize,labelfont=sf,textfont=sf]{subfig}
\usepackage{textcomp}
\usepackage{stfloats}
\usepackage{url}
\usepackage{verbatim}
\usepackage{graphicx}
\usepackage{cite}
\usepackage{lipsum}     
\usepackage{booktabs} 
\usepackage{tabularx} 
\usepackage{wrapfig} 
\usepackage{multirow}
\usepackage[section]{placeins} 
\usepackage{balance}
\ifCLASSINFOpdf
\else
\fi
\begin{document}
%
\title{ManiVID-3D: Generalizable View-Invariant Reinforcement Learning for Robotic Manipulation via Disentangled 3D Representations}

\author{Zheng Li$^{1}$, Pei Qu$^{1}$, Yufei Jia$^{2}$, Shihui Zhou$^{1}$, Haizhou Ge$^{2}$, Jiahang Cao$^{3}$,\\
Jinni Zhou$^{1}$, Guyue Zhou$^{2}$, Jun Ma$^{1}$,~\IEEEmembership{Senior Member,~IEEE}
\thanks{Manuscript received: September 12, 2025; Revised January 1 2026; Accepted January 26, 2026.}
\thanks{This paper was recommended for publication by Editor Jens Kober upon evaluation of the Associate Editor and Reviewers' comments.
This work was supported by the Guangdong Provincial Educational Science Planning Project (Grant No. 2025GXJK0629), Guangzhou Higher Education Teaching Research and Reform Project (Grant No. 2024YBJG094), and Guangzhou Municipal Educational Science Planning Project (Grant No. 202316577). \textit{(Zheng Li, Pei Qu, and Yufei Jia contributed equally to this work) (Corresponding author: Jinni Zhou)}} 
\thanks{$^{1}$Zheng Li, Pei Qu, Shihui Zhou,  Jinni Zhou, and Jun Ma are with The Hong Kong University of Science and Technology (Guangzhou), Guangzhou 511453, China (e-mail: zli514@connect.hkust-gz.edu.cn).}%
\thanks{$^{2}$Yufei Jia, Haizhou Ge, and Guyue Zhou are with Tsinghua University, Beijing 100084, China (e-mail: jyf23@mails.tsinghua.edu.cn).}
\thanks{$^{3}$Jiahang Cao is with The University of Hong Kong, Hong Kong SAR, China (e-mail: jiahang@connect.hku.hk).}
\thanks{The project page can be found at https://zheng-joe-lee.github.io/manivid3d/.}
\thanks{Digital Object Identifier (DOI): see top of this page.}
}

%
%

\markboth{IEEE Robotics and Automation Letters. Preprint Version. Accepted JANUARY, 2026}
{Li \MakeLowercase{\textit{et al.}}: ManiVID-3D: Generalizable View-Invariant Reinforcement Learning for Robotic Manipulation} 

%



\maketitle



\begin{abstract}

Deploying visual reinforcement learning (RL) policies in real-world manipulation is often hindered by camera viewpoint changes. A policy trained from a fixed front-facing camera may fail when the camera is shifted—an unavoidable situation in real-world settings where sensor placement is hard to manage appropriately. Existing methods often rely on precise camera calibration or struggle with large perspective changes. To address these limitations, we propose ManiVID-3D, a novel 3D RL architecture designed for robotic manipulation, which learns view-invariant representations through self-supervised disentangled feature learning. The framework incorporates ViewNet, a lightweight yet effective module that automatically aligns point cloud observations from arbitrary viewpoints into a unified spatial coordinate system without the need for extrinsic calibration. Additionally, we develop an efficient GPU-accelerated batch rendering module capable of processing over 5000 frames per second, enabling large-scale training for 3D visual RL at unprecedented speeds. Extensive evaluation across 10 simulated and 5 real-world tasks demonstrates that our approach achieves a 40.6\% higher success rate than state-of-the-art methods under viewpoint variations while using 80\% fewer parameters. The system's robustness to severe perspective changes and strong sim-to-real performance highlight the effectiveness of learning geometrically consistent representations for scalable robotic manipulation in unstructured environments. 

\end{abstract}

\IEEEpeerreviewmaketitle
\section{Introduction} \label{sec 1}

\IEEEPARstart{V}{isual} reinforcement learning (RL) has shown great promise for robotic manipulation, enabling end-to-end learning of complex visuomotor skills like grasping \cite{lee2025grasping, chen2021deep}, dexterous manipulation \cite{jia2025multi, xu2025hierarchical}, and bimanual manipulation \cite{amadio2019exploiting} directly from visual inputs. Among these, 3D vision-based RL offers key advantages over 2D approaches by providing inherent geometric structure, improving spatial reasoning, and enabling more precise action prediction are all critical for real-world manipulation~\cite{chen2021deep, xu2025hierarchical, amadio2019exploiting, peri2024point, ze2023visual_RL3D}.


  
  
  

\begin{figure}[!t] 
\raggedleft 
\includegraphics[width=0.48\textwidth]{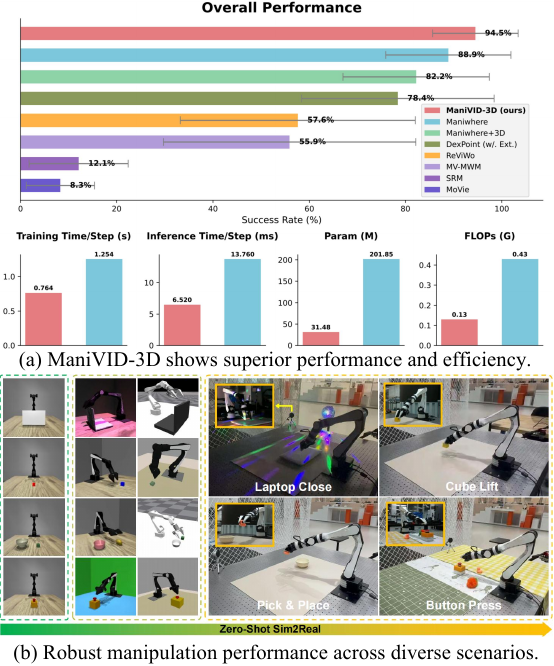}
\caption{\textbf{ManiVID-3D.} Our method achieves robust multi-domain generalization for manipulation tasks with superior viewpoint adaptation and sim-to-real transferability, while significantly reducing computational costs.}
\label{fig:1}
\end{figure}

However, a primary challenge in deploying a trained policy in real-world scenarios lies in its generalization capability~\cite{RoboTwin, discoverse}, particularly across viewpoint variations, which remains one of the most significant yet underexplored aspects of cross-domain robustness~\cite{mvmwm, robouniview}. For instance, a robot trained to pick up a cup from a tabletop may succeed consistently when observed from a designated training camera, but fail when the same scene is viewed from a slightly shifted angle—an unavoidable situation in cluttered households or dynamic industrial settings where cameras cannot always be rigidly fixed. Unlike other nuisance factors such as lighting~\cite{qu2025fast, RoboTwin2.0} or texture~\cite{RoboTwin2.0, chen2025g3flow}, viewpoint shifts cause drastic geometric distortions in visual observations, making it particularly hard for policies to generalize~\cite{robouniview, ManiWhere, vista}. This gap underscores the need for learning view-invariant representations, enabling robots to reliably perceive and understand 3D environments across diverse sensor placements. 

While recent efforts have attempted to address the challenge of viewpoint generalization, most state-of-the-art methods rely on 2D visual inputs. For instance, MV-MWM \cite{mvmwm} employs a multi-view masked autoencoder (MAE) \cite{mae} to reconstruct viewpoint-robust image features.
However, such 2D representations struggle to capture the essential 3D structural priors required for robust cross-view generalization, resulting in suboptimal performance. Although depth-aware methods like Maniwhere \cite{ManiWhere} demonstrate notable improvements by incorporating partial geometric cues, their reliance on depth maps, which inherently lack complete geometric information and are susceptible to occlusion, leads to substantial performance degradation under large viewpoint variations. RoboUniView~\cite{robouniview} adopts point cloud inputs to learn 3D unified multi-view representations, but it depends on precise camera extrinsic calibration, which severely constrains their practicality in dynamic real-world settings.

To address the limitations of existing approaches, we propose \textbf{ManiVID-3D}: A \textbf{V}iew-\textbf{I}nvariant Visual RL Architecture based on \textbf{D}isentangled 3D Representations for robotic \textbf{mani}pulation. By integrating supervised view alignment with self-supervised feature disentanglement, our method successfully learns 3D visual representations that are robust to large viewpoint variations, overcoming the reliance on strict camera calibration. For large-scale training, we implement GPU-accelerated parallel point cloud generation via our Efficient Batch Simulation with Rendering module, achieving high throughput across environments. To validate the effectiveness of our approach, we conduct extensive experiments across 10 manipulation tasks, involving 3 types of robotic arms and 3 gripper end-effectors, with evaluations in both simulation and real-world environments. As shown in Fig. \ref{fig:1}, ManiVID-3D significantly outperforms most state-of-the-art methods with exceptional sim-to-real transfer capability. Compared to the strongest baseline, Maniwhere~\cite{ManiWhere}, our architecture achieves superior cross-view generalization while reducing parameters by over 80\%. Furthermore, we investigate ManiVID-3D’s performance on other scene generalization types beyond viewpoint variation to provide a comprehensive analysis of its robustness.

Our core contributions can be summarized as follows:
\begin{itemize}

\item We propose ManiVID-3D, a novel 3D visual RL architecture using disentangled view-invariant representations that exhibits strong  robustness toward extreme viewpoint changes during manipulation tasks.
\item We introduce ViewNet, a plug-and-play multi-view alignment module that unifies point cloud coordinates across viewpoints, eliminating dependency on strict camera calibration.
\item We develop a GPU-accelerated batch rendering system that enables large-scale parallel simulation with high-quality point cloud observations at unprecedented speeds of thousands of frames per second.
\item Through extensive experiments, we demonstrate that ManiVID-3D achieves significantly superior success rate than state-of-the-art baselines under extreme viewpoint changes with substantially fewer parameters, while maintaining superior sim-to-real transferability.

\end{itemize}
\section{Related Work}\label{sec 2}

\subsection{View Generalization in Robot Manipulation}
Enabling robotic policies to generalize to unseen camera viewpoints remains a significant challenge in visual reinforcement learning. Current approaches primarily employ two strategies, each with notable limitations. Data-centric methods such as domain randomization \cite{discoverse, RoboTwin2.0, chen2025g3flow, DP3, iDP3}, domain adaptation \cite{MoVie}, and data augmentation \cite{vista, ma2022comprehensive_survey, roviaug, SRM} aim to expose policies to sufficient visual variability during training, but face a fundamental trade-off: insufficient augmentation yields limited generalization, while excessive perturbation often leads to training instability or divergence \cite{ManiWhere, ma2022comprehensive_survey}. Alternatively, representation learning methods like ManiWhere \cite{ManiWhere} and ReViWo \cite{reviwo} attempt to learn view-invariant features through auxiliary objectives such as reconstruction losses \cite{mvmwm, robouniview, reviwo} or contrastive losses \cite{ManiWhere, reviwo}. While effective to some degree, these approaches are inherently constrained by their reliance on 2D visual inputs, which lack explicit 3D geometric understanding and consequently degrade under large viewpoint changes. Exciting 3D-aware methods (e.g., RoboUniView \cite{robouniview}, RVT~\cite{rvt}) typically depend on precise camera calibration, severely limiting their practical applicability in real-world scenarios where camera parameters may be unknown or variable. These limitations motivate our work in developing a calibration-free 3D representation that explicitly preserves geometric relationships across views, while avoiding the instability pitfalls of strong data augmentation.



\subsection{Disentanglement with Contrastive Learning}
Disentangled representation learning aims to align individual dimensions or sub-vectors of a latent space with semantically meaningful factors, enhancing model interpretability, controllability, and generalization capabilities~\cite{wang2024disentangled}. Among various unsupervised/self-supervised disentanglement approaches \cite{Lvae, jun2025disentangling}, contrastive learning has emerged as a particularly promising paradigm due to its intuitive formulation and inherent capability to separate latent factors \cite{mo2023representation, matthes2023towards}. 
By mapping observed data into an embedding space where related samples (e.g., different augmented views of the same instance) are pulled together while unrelated pairs are pushed apart \cite{infonce}, contrastive learning naturally facilitates the decomposition of disentangled representations. 
Recent advances in visuomotor policy learning have adopted this framework to decouple multi-view representations, such as separating first-person and third-person visual features~\cite{disdp, almuzairee2025merging, dunion2024multi}. Building upon these works, we extend contrastive disentanglement to single-view policy learning, enabling explicit separation of view-invariant and view-dependent features for cross-view generalization.

\section{Preliminaries}\label{sec 3}

\begin{figure*}[t] 
\vspace{5pt}
\centering
\includegraphics[width=0.83\textwidth]{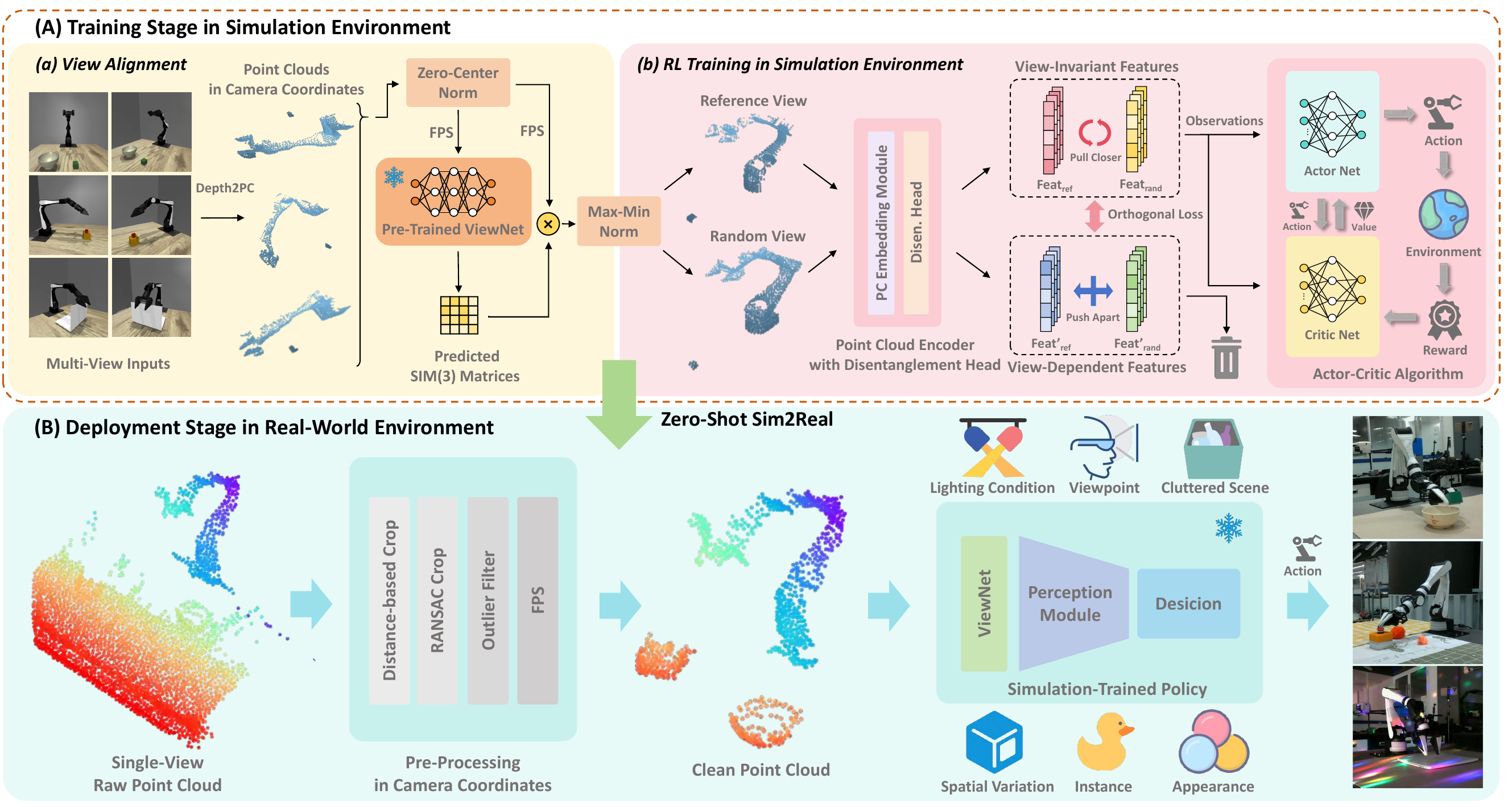} 
\caption{\textbf{Overview of ManiVID-3D.} (A) In the training phase, our method consists of two key components: (a) Pretrained ViewNet aligns arbitrary-viewpoint clouds collected in simulation to a unified frame without extrinsic calibration; (b) A disentanglement encoder extracts view-invariant features that are used to train manipulation policies with strong cross-view generalization. (B) In the deployment phase, we introduce a multi-stage processing pipeline specifically designed for camera-coordinate point clouds to bridge the sim-to-real domain gap, enabling zero-shot transfer to real-world deployment.}
\label{fig:model}
\end{figure*}

\subsection{Visual Reinforcement Learning}
Visual RL formulates control from high-dimensional observations (e.g., images or point clouds) as a Partially Observable Markov Decision Process (POMDP) \cite{smallwood1973optimal}, defined by $\langle\mathcal{S}, \mathcal{A}, \mathcal{O}, T ,R, \gamma\rangle$.
In this work, observations $\mathcal{O}(s)\in\mathbb{R}^{N\times K\times C}$ consist of $N$ consecutive point cloud frames with $K$ points per frame and $C$-dimensional features, including 3D coordinates and optional attributes such as RGB.
The agent aims to learn an optimal policy $\pi^{*}:\mathcal{O}\to\Delta(\mathcal{A})$ that maximizes the expected discounted return
$J(\pi)=\mathbb{E}_{\pi, T}\left[\sum_{t=0}^\infty\gamma^tR(s_t,a_t)\right]$.

\subsection{Contrastive Learning}
In this work, contrastive learning is employed to learn disentangled view-dependent and view-invariant representations of point cloud in a self-supervised manner. In contrastive learning, for a given query embedding $q$ (also known as an anchor), the objective is to enhance the similarity between $q$ and a positive sample  $k^+$, while simultaneously reducing its similarity with all negative samples $\{k_{i}^{-}\}_{i=0}^{N-1}$. In recent years, the most widely used objective function for this purpose is InfoNCE \cite{infonce}. Given a similarity function and temperature coefficient $\tau$, InfoNCE is defined as follows:
\begin{equation} \label{infonce}
    \begin{aligned}
        &\mathcal{L}^{\text{InfoNCE}}(q,k^+,\{k_{i}^{-}\}_{i=0}^{N-1})\\
        &=-\log\frac{\exp(q^Tk^+/\tau)}{\exp(q^Tk^+/\tau)+\sum_{i=0}^{N-1}\exp(q^Tk^-_i/\tau)}.
    \end{aligned}
\end{equation}
Here, the similarity function is defined as the dot product $q^Tk$, which is a widely adopted practice in contrastive learning \cite{chen2020simple(Contrastive_dot)}. 
\section{Method}\label{sec 4}

In this section, we introduce \textbf{ManiVID-3D}, a  novel view-invariant manipulation policy learning framework for 3D visual reinforcement learning. As shown in Fig.~\ref{fig:model}, the framework consists of two main stages. First, a pretrained ViewNet aligns point clouds from arbitrary viewpoints to a unified reference frame without camera extrinsic calibration. Second, a disentanglement encoder with a tailored objective extracts view-invariant representations from the normalized point clouds, which are used to train visuomotor policies with strong cross-view generalization. To enable large-scale training, an efficient batch simulation with GPU-accelerated rendering supports high-throughput point-cloud generation. Additionally, a multi-stage pipeline optimized for camera-coordinate point clouds is introduced to facilitate zero-shot sim-to-real transfer.
In the remainder of this section, we provide a comprehensive exposition of the methodological details of our approach.

\subsection{Learning Cross-View Representation with Point Clouds}

Point cloud representation has demonstrated superior performance in robot learning compared to alternative 3D representations \cite{DP3, zhu2024point, peri2024point}. In ManiVID-3D, we employ single-view sparse point clouds as the observation space for actor-critic algorithm. During simulation training, our system acquires 240$\times$240 depth images from two cameras: one fixed and one randomly positioned. These depth measurements are transformed into camera-frame points using intrinsics. While we utilize dual-camera data collection during training to facilitate contrastive learning, we emphasize that our framework requires only a single camera during testing, both in simulation and real-world deployment.

\textbf{Multi-stage point cloud processing.}
Processing camera-frame point clouds presents unique challenges, particularly in efficient cropping and normalization. Unlike iDP3's computationally expensive full-scene inputs \cite{iDP3}, our framework introduces a multi-stage preprocessing pipeline specifically optimized for robotic manipulation: (1) Workspace cropping using conservative depth bounds, (2) RANSAC-based plane removal for redundant surfaces (e.g., tables, walls), (3) Statistical outlier filtering for real-world data robustness, and (4) Truncated min-max normalization using percentile thresholds to mitigate noise effects. The processed point cloud then undergoes farthest point sampling (FPS), maintaining geometric fidelity while ensuring computational efficiency.

\textbf{Implementation details of point cloud preprocessing.}
Notably, all preprocessing hyperparameters are fixed and shared across all tasks, objects, and scenes, with no per-task or per-scene tuning. The parameters are selected once based on the depth sensor characteristics and the robot manipulation workspace, using a small set of preliminary point clouds outside the evaluation scenes. Specifically, we apply: (1) depth cropping at 2.0 m, (2) RANSAC plane removal with a 0.02 m inlier threshold, a 3-point plane model, and 1000 iterations, (3) statistical outlier removal with 30 neighbors and a standard deviation ratio of 2.0, and (4) min–max normalization using the 5th–95th percentiles. These settings are stable across scenes and tasks, and moderate variations do not significantly affect performance, indicating that no scene-specific tuning is required.


\textbf{Data augmentation.} To enhance the robustness and generalization capabilities of our learned representations, we implement a comprehensive data augmentation strategy specifically designed for point cloud manipulation tasks. Our augmentation pipeline includes: (1) random geometric transformations including rotation, translation, and scaling within physically plausible ranges, (2) point dropout and subsampling to simulate partial observations, and (3) Gaussian noise injection to model sensor uncertainty.

\subsection{Contrastive Learning for Feature Disentanglement}

\textbf{Point cloud encoder with disentanglement head.} 
Inspired by DP3~\cite{DP3}, we adopt a lightweight MLP architecture with key modifications for effective representation disentanglement. Specifically, We implement this through a novel dual-head design: the third MLP layer branches into two parallel, structurally identical heads to extract view-invariant features for task execution and view-dependent features for viewpoint modeling. This design reduces parameters by 80\% compared to the RGB-D encoder in Maniwhere~\cite{ManiWhere}, while preserving effective disentanglement, as shown in our qualitative analysis (Sec.~\ref{sec QA}).

\textbf{View-invariant disentanglement objective.} 
We propose a novel view-invariant representation learning objective that explicitly decouples view-invariant and view-dependent features for robust 3D perception. At each timestep $k$, the environment provides paired point clouds $(p_k^{\text{ref}}, p_k^{\text{rand}})$ captured from a fixed reference and a randomly sampled viewpoint. The encoder $f_{\theta,\phi,\psi}$ shares parameters $\theta$ in the first two MLP layers and branches into separate heads for view-invariant ($\phi$) and view-dependent ($\psi$) representations.

For view-invariant features, we enforce cross-view consistency using an InfoNCE contrastive loss, where the reference-view embedding at timestep $k$ serves as the query, the corresponding random-view embedding as the positive, and random-view embeddings from other timesteps in the same batch $B$ as negatives:
\begin{align}
    \mathcal{L}_{\text{inv}}=\mathcal{L}^{\text{InfoNCE}}(f_{\phi}(p_k^{\text{ref}}), f_{\phi}(p_k^{\text{rand}}), \{f_{\phi}(p_i^{\text{rand}})\}_{i\in B}^{i \neq k}).
\end{align}

Conversely, for viewpoint-dependent representations, we enforce feature divergence across different camera perspectives. While maintaining the same query embedding (reference view at timestep $k$), we define positive samples as the reference view's embeddings from other timesteps ($t \neq k$), while negative samples comprise all embeddings captured from the random viewpoints in the batch samples $B$:
\begin{align}
    \mathcal{L}_{\text{dep}}=\mathcal{L}^{\text{InfoNCE}}(f_{\psi}(p_k^{\text{ref}}), f_{\psi}(p_{t,t \neq k}^{\text{ref}}), \{f_{\psi}(p_i^{\text{rand}})\}_{i\in B}).
\end{align}

To ensure enough decoupling between these two representations, we introduce an orthogonal loss function that explicitly minimizes their mutual interference. The loss is formulated as:
\begin{align}
    \mathcal{L}_{\text{orth}}=\frac{||f_{\phi}(p_k^{\text{ref}})^Tf_{\psi}(p_k^{\text{ref}})||+||f_{\phi}(p_k^{\text{rand}})^Tf_{\psi}(p_k^{\text{rand}})||}{2}.
\end{align}

To prioritize viewpoint-invariant feature learning while maintaining RL training stability, we introduce a time-dependent scaling coefficient $\beta$ (initially $\beta\approx0$) that linearly increases during training. This curriculum strategy initially suppresses the viewpoint-dependent loss term, allowing the encoder to first establish robust viewpoint-invariant representations. The complete training objective thus becomes:
\begin{align}
    \mathcal{L}_{\text{VID}}=\mathcal{L}_{\text{inv}}+\beta(t)(\mathcal{L}_{\text{dep}}+\lambda\mathcal{L}_{\text{orth}}),
\end{align}
where $\lambda$ controls the orthogonality constraint strength. We use a linear warm-up schedule for the scaling coefficient $\beta$:
\begin{align}
    \beta(t)=\text{max}(\beta_{\text{min}}, \text{min}(\beta_{\text{max}}, (t-t_0)/T)),
\end{align}
where $t$ is the training step, $t_0$ is the contrastive loss start step, $T=100k$ is the warm-up duration. Thus, $\beta$ increases linearly from $\beta_{\text{min}}=0.0$ to  $\beta_{\text{max}}=1.0$ and remains at $\beta_{\text{max}}$ afterwards. In Sec.~\ref{sec abla}, we further conduct a quantitative analysis on $\beta$ and report its impact on performance.

\begin{table*}[t]
  \vspace{5pt}
  \centering
  \caption{\textbf{Performance comparison across different viewpoints in simulation.} ManiVID-3D outperforms representative baselines by \textbf{+40.6\%} success rate on average across 10 diverse manipulation tasks under randomized camera viewpoints.}
  \label{tab:1 main}
  \resizebox{\textwidth}{!}{ 
      \begin{tabular}{lccccccccccc}
        \toprule
        & \multicolumn{5}{c}{AIRBOT Play} & \multicolumn{3}{c}{UR5} & \multicolumn{2}{c}{Franka} &   \\
        \cmidrule(lr){2-6} \cmidrule(lr){7-9} \cmidrule(lr){10-11}
        Method & Reach & Cube Lift & Pick \& Place & Button Press & Laptop Close & Drawer Open & Reach Dex & Button Press Dex & Pick \& Place Dex & Hand Over Dual & Average \\
        \midrule
        Maniwhere \cite{ManiWhere} & 97.2$\pm$2.3 & 86.4$\pm$5.6 & 88.4$\pm$7.8 & 96.4$\pm$1.7 & 82.8$\pm$16.2 & 76.4$\pm$10.6 & 96.8$\pm$2.3 & 96.0$\pm$1.8 & 76.0$\pm$8.6 & 93.2$\pm$5.5 & 88.9 \\
        Maniwhere + 3D & 90.4$\pm$6.8 & 78.4$\pm$9.2 & 69.6$\pm$7.5 & 93.2$\pm$5.6 & 86.4$\pm$8.3 & 68.8$\pm$9.2 & 94.0$\pm$8.4 & 90.4$\pm$6.5 & 65.2$\pm$12.6 & 85.6$\pm$8.7 & 82.2 \\
        DexPoint (w/ Ext.) \cite{dexpoint} & 91.6$\pm$4.7 & 80.8$\pm$7.2 & 63.2$\pm$9.4 & 89.6$\pm$8.4 & 78.4$\pm$12.6 & 71.6$\pm$8.5 & 90.8$\pm$9.0 & 83.6$\pm$11.6 & 56.4$\pm$17.6 & 78.0$\pm$7.5 & 78.4 \\
        ReViWo \cite{reviwo} & 71.2$\pm$18.4 & 63.2$\pm$15.6 & 44.4$\pm$19.7 & 69.6$\pm$15.3 & 58.0$\pm$16.8 & 56.6$\pm$21.7 & 70.4$\pm$12.6 & 64.8$\pm$15.3 & 36.2$\pm$25.5 & 41.6$\pm$13.6 & 57.6 \\
        MV-MWM \cite{mvmwm}  & 76.8$\pm$15.2 & 59.6$\pm$18.4 & 30.8$\pm$23.5 & 61.2$\pm$23.5 & 66.8$\pm$17.2 & 47.2$\pm$23.5 & 76.4$\pm$9.3 & 75.2$\pm$12.6 & 29.6$\pm$24.6 & 35.2$\pm$18.5 & 55.9 \\
        SRM \cite{SRM} & 20.4$\pm$3.2 & 12.4$\pm$5.8 & 2.0$\pm$1.2 & 16.4$\pm$5.2 & 10.4$\pm$3.4 & 4.8$\pm$2.0 & 19.6$\pm$6.2 & 17.2$\pm$5.2 & 5.2$\pm$4.8 & 12.4$\pm$7.8 & 12.1 \\
        MoVie \cite{MoVie} & 14.4$\pm$2.8 & 11.2$\pm$3.2 & 1.2$\pm$2.8 & 10.8$\pm$5.3 & 4.8$\pm$3.6 & 5.2$\pm$3.2 & 12.8$\pm$4.5 & 9.6$\pm$4.8 & 1.6$\pm$2.8 & 11.2$\pm$5.6 & 8.3 \\
        \midrule
        \textbf{ManiVID-3D (ours)} & \textbf{99.2$\pm$0.8} & \textbf{92.0$\pm$7.5}  & \textbf{91.2$\pm$5.2} & \textbf{98.8$\pm$1.2} & \textbf{95.6$\pm$3.7} & \textbf{85.6$\pm$8.8} & \textbf{99.6$\pm$0.4} & \textbf{98.4$\pm$1.5} & \textbf{88.8$\pm$5.2} & \textbf{95.6$\pm$2.5} & \textbf{94.5} \\
        \bottomrule
      \end{tabular}
    }
\end{table*}

\subsection{ViewNet}
To remove reliance on accurate camera extrinsics, we introduce ViewNet, a pretrained module that aligns point clouds from arbitrary viewpoints to a unified reference frame. ViewNet uses a PointNet++~\cite{pointnet++} backbone with a regression head that predicts SIM(3) transformation parameters, which are applied to input points via spatial warping. Exploiting simulator-enabled data efficiency, ViewNet is trained on synthetic triplets consisting of the raw input $p_{\text{org}}$, its world-coordinate version $p_{\text{world}}$, and a fixed-reference point cloud $p_{\text{ref}}$. The resulting composite loss is defined as:
\begin{align}
    \mathcal{L}_{\text{ViewNet}}=\frac{\mathcal{L}_{\text{MSE}}(p_{\text{org}}, p_{\text{world}})+\mathcal{L}_{\text{Chamfer}}(p_{\text{org}}, p_{\text{ref}})}{2},
\end{align}
where chamfer distance between two point sets $A$ and $B$ is:
\begin{align}
&\mathcal{L}_{\text{Chamfer}}(A,B) \nonumber \\
&= \frac{1}{|A|}\sum_{a \in A} \min_{b \in B} \|a - b\|_2^2 + \frac{1}{|B|}\sum_{b \in B} \min_{a \in A} \|a - b\|_2^2.
\end{align}

Once trained, ViewNet operates as a plug-and-play module across diverse tasks, enabling agents to comprehend 3D environments without reliance on extrinsic calibration while demonstrating exceptional sim-to-real transferability.  To ensure robustness under realistic partial observability, ViewNet does not assume the entire robot or object is visible. During training, we apply point-cloud occlusion augmentation: about 30\% of training samples include spatially contiguous block occlusions, where 20–40\% of points are masked out within each block. This strategy improves robustness in occluded scenarios, with detailed qualitative analysis in Sec.~\ref{sec QA}.

\begin{figure}[!t] 
\raggedleft 
\includegraphics[width=0.48\textwidth]{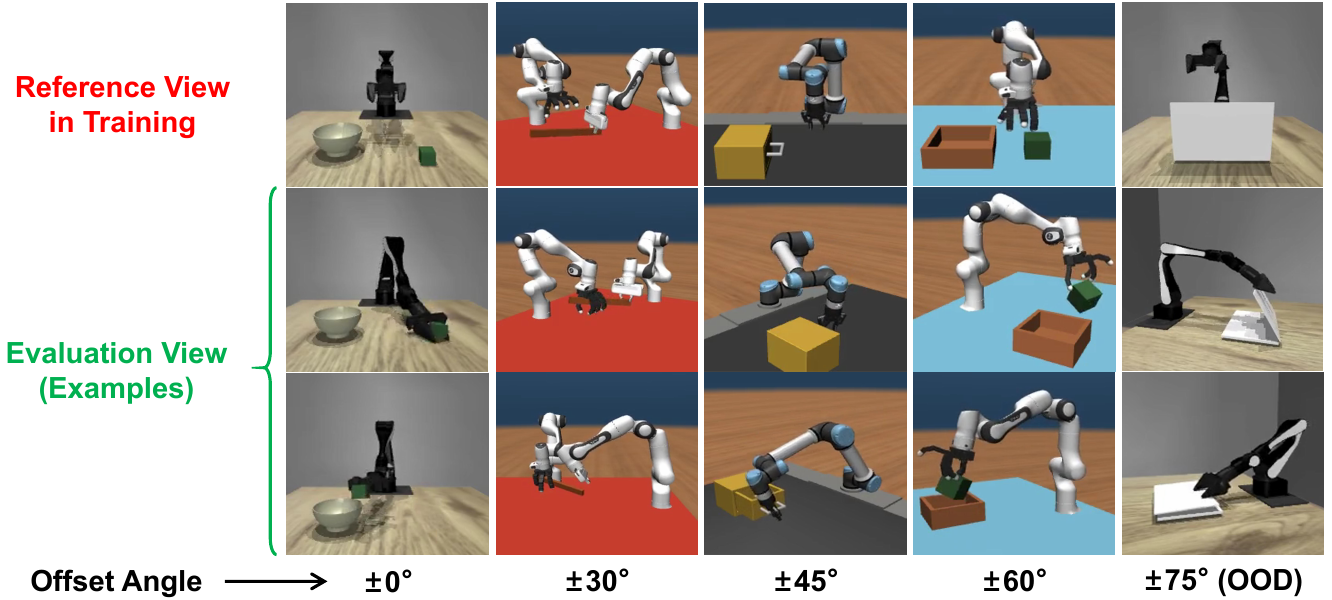}
\caption{\textbf{Simulation snapshots.} Reference view and evaluation view at different angular offsets for selected tasks.}
\label{fig:3}
\end{figure}

\subsection{Efficient Batch Simulation with Rendering}

Contemporary RL for robotic manipulation is bottlenecked by visual perception because high-fidelity rendering during training is costly. While parallel simulators such as Isaac Gym can train thousands of robots concurrently, they often avoid rich vision by using only proprioception or low-resolution height maps, which limits complex scene understanding. We address this by building on ManiSkill3’s parallel architecture~\cite{maniskill3} and introducing a GPU-accelerated batch renderer that generates 3D observations at scale. It delivers thousands of FPS with high-quality 3D geometry, enabling on-policy RL for thousands of agents in parallel.


\section{Experiments}

\textbf{Tasks.} 
In simulation, we adopt a Maniwhere-inspired benchmark comprising 10 progressively challenging tasks in the MuJoCo engine~\cite{mujoco} (the parallel training implementation is built upon Sapien~\cite{xiang2020sapien}). The tasks cover three robotic arms (AIRBOT Play, UR5, and Franka) and diverse manipulation settings, including single-arm, dual-arm, and dexterous manipulation. Task details are summarized in Table~\ref{tab:1 main}, with representative tasks and camera configurations shown in Fig.~\ref{fig:3}. For real-world evaluation, we select five AIRBOT Play tasks for hardware experiments.

\textbf{Baselines.} 
Our baselines include state-of-the-art visual RL algorithms representing key technical approaches: domain adaptation (MoVie \cite{MoVie}), data augmentation (SRM \cite{SRM}), auto-encoder pretraining (MV-MWM \cite{mvmwm}, ReViWo \cite{reviwo}), and self-supervised contrastive learning (Maniwhere \cite{ManiWhere}). To verify our method's advantages are not solely due to point cloud inputs, we include a 3D Maniwhere variant with a SIM(3)-capable STN \cite{stn,3dstn}. 
Furthermore, we add DexPoint~\cite{dexpoint}, a representative point-cloud-based 3D visual RL method, as a baseline. For a fair comparison under cross-view evaluation, DexPoint is also provided with camera extrinsics, which our method does not require.

\textbf{Robot setup.} In simulation, our experiments employ four robotic manipulator configurations: (1) the AIRBOT Play arm with AIRBOT Gripper 2, (2) the UR5 arm paired with a Robotiq gripper, (3) the UR5 arm integrated with an Allegro Hand, and (4) the Franka Emika arm coupled with an Allegro Hand. For real-world validation, we deploy the AIRBOT Play arm equipped with the AIRBOT Gripper 2 as the hardware platform. Depth data is captured via a RealSense L515 camera and converted into point clouds in the camera coordinate frame to serve as input for our system.

\textbf{Evaluation metrics.} The primary metric for evaluating our approach is the task success rate achieved by trained models during the testing phase. For each task and model configuration, we conduct evaluations across 5 seeds to ensure statistical reliability, with the final performance reported as the mean success rate over these trials.

\subsection{Simulation Experiments}\label{sec simu}

\textbf{Performance evaluation against the state-of-the-arts.} 
We extensively evaluate ManiVID-3D against multiple strong baselines on 10 diverse tasks. Each task is tested over 50 episodes with randomized viewpoints, including yaw variations of $\pm\text{60}^\circ$, pitch variations of $\pm\text{7.5}^\circ$, and camera distances scaled to 0.9–1.1× of the reference. As shown in Table~\ref{tab:1 main}, ManiVID-3D consistently outperforms all existing SOTA methods across all tasks. Notably, directly adapting Maniwhere to point cloud inputs with its original contrastive loss fails to sufficiently constrain the 3D-STN transformation, resulting in inferior performance even compared to the original Maniwhere.
Our method also consistently outperforms DexPoint, a strong point-cloud-based RL baseline, even when DexPoint is provided with camera extrinsics to improve cross-view generalization. Overall, our method achieves an average 40.6\% improvement in success rate over the baselines and a 5.6\% gain over Maniwhere.

\begin{figure}[!t] 
\vspace{5pt}
\raggedleft 
\includegraphics[width=0.48\textwidth]{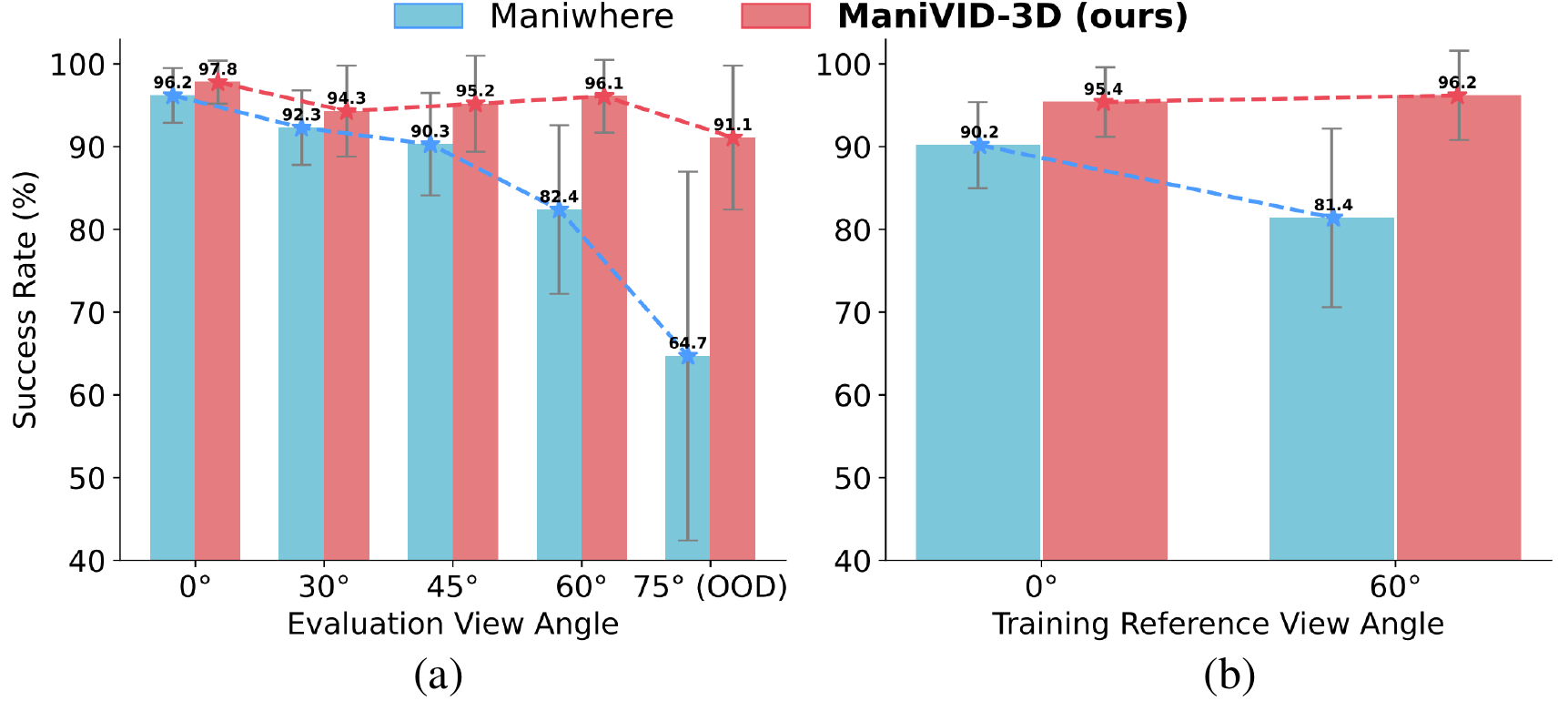}
\caption{\textbf{Robustness to (a) viewpoint variation and (b) reference viewpoint selection.}  ManiVID-3D maintains consistently strong performance across varying degrees of view offsets and different reference viewpoint choices, whereas Maniwhere exhibits a clear performance degradation trend.}
\label{fig:4}
\end{figure}

\textbf{Robustness analysis under extreme viewpoint variations.} 
Our experiments reveal two key limitations of RGB-D methods such as Maniwhere: (1) severe performance degradation under large viewpoint shifts ($>\text{45}^\circ$), likely due to the inability of STN modules to align 3D structure from depth maps, and (2) sensitivity to the training reference viewpoint, with an 8.8\% success-rate drop when deviating from frontal views. We evaluate Maniwhere and ManiVID-3D on the five AIRBOT Play tasks under increasing yaw offsets ($\pm30^\circ$ to $\pm75^\circ$; training range $[\text{-70}^\circ,\text{70}^\circ]$) and different training viewpoints (frontal vs. $\text{60}^\circ$ lateral). As shown in Fig.~\ref{fig:4}, Maniwhere exhibits accelerating performance decay (31.5\% at $\pm\text{75}^\circ$ vs. 3.9\% at $\pm\text{30}^\circ$), whereas ManiVID-3D remains stable ($<$6.7\% variance) and even improves performance by 0.8\% with lateral reference views, likely due to reduced occlusions in lateral viewpoints. These results demonstrate the stronger view invariance of our method.

\begin{figure}[!t] 
\vspace{5pt}
\raggedleft 
\includegraphics[width=0.48\textwidth]{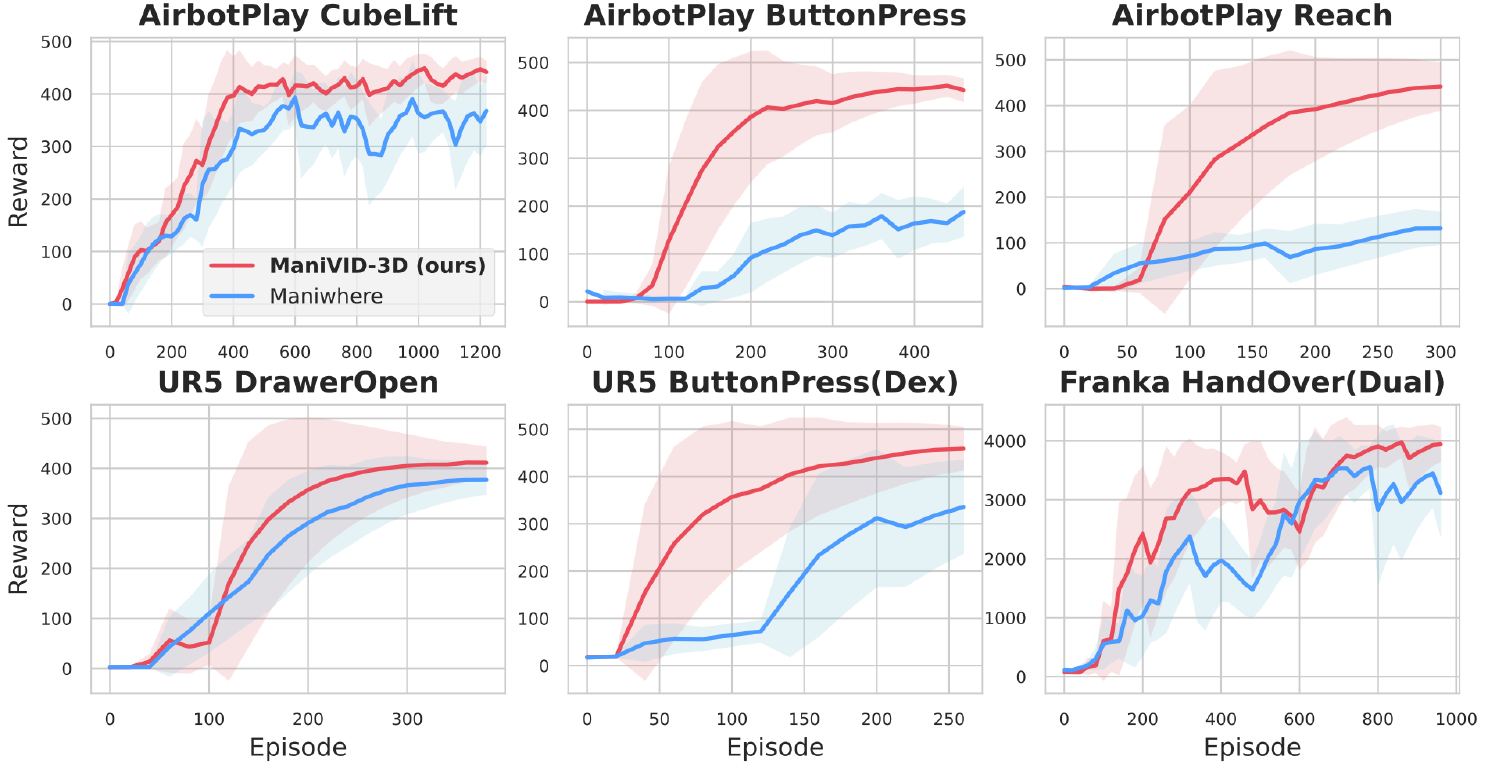}
\caption{\textbf{RL training curves.} ManiVID-3D shows superior convergence to Maniwhere in most tasks.}
\label{fig:5}
\end{figure}

\textbf{Efficiency analysis.} Beyond higher success rates, our method is substantially more computationally efficient. Benchmarking highlights strong efficiency gains: on an RTX 3090, our method reduces training time by 39.1\% while converging faster and achieving higher final rewards than Maniwhere (Fig. \ref{fig:5}). At inference, it cuts latency by 52.6\% with 84.4\% fewer parameters and 69.8\% fewer FLOPs, enabling real-time use. We further speed up learning with an Efficient Batch Simulation with Rendering module that co-optimizes parallel rendering and physics. On an RTX 4090 with 512 environments, 2 cameras, and 32k points each, we reach 10.0 FPS/env (total 5127.7 FPS), far exceeding MuJoCo’s ~30 FPS serial rendering, demonstrating reduced overhead and improved overall efficiency.

\textbf{Comparison with imitation learning (IL).}
To further validate the effectiveness of our approach, we conduct comparisons with state-of-the-art 3D Imitation Learning  algorithms. 
\begin{wraptable}{r}{0.2875\textwidth} 
\centering\small
\caption{\textbf{Comparison results with imitation learning.}}
\begin{tabular}{l|c}
\toprule
Method & Success Rate\\
\midrule
DP3 \cite{DP3} & 18.0$\pm$5.0 \\
ManiVID-3D (IL) & 78.4$\pm$5.6 \\
\textbf{ManiVID-3D (RL)} & \textbf{92.0$\pm$7.5} \\
\bottomrule
\end{tabular}

\label{tab:il}
\end{wraptable}
Specifically, we evaluate performance on the cube lift task, collecting 100 demonstration trajectories from multiple viewpoints for training of 3D Diffusion Policy (DP3) \cite{DP3}. 
Additionally, we adapt ManiVID-3D to an IL training paradigm and include it in the comparison. As shown in Table~\ref{tab:il}, ManiVID-3D substantially outperforms DP3 under both RL and IL training, indicating that the advantage of our viewpoint-invariant 3D representation is not tied to a specific training paradigm.

\begin{table*}[htbp]
  \vspace{5pt}
  \centering
  \caption{\textbf{Real-world experiments results.} ManiVID-3D demonstrates superior performance over Maniwhere in all tasks across various complex real-world scenarios and viewpoint changes.}
  \label{tab:2real}
  \resizebox{\textwidth}{!}{ 
    \begin{tabular}{l ccccc cccc c}
      \toprule
        & 
      \multicolumn{5}{c}{Viewpoint Variation} & 
      Appearance & Instance & Lighting Condition & Cluttered Scene & \\
      \cmidrule(lr){2-6} \cmidrule(lr){7-10}
       Method & Reach & Cube Lift & Pick \& Place & Laptop Close & Button Press & \multicolumn{4}{c}{Cube Lift} & Total\\
      \midrule
    
      Maniwhere \cite{ManiWhere} & 
      16/20 & 11/20 & 8/20 & 13/20 & 14/20 & 
      5/10 & 3/10 & 5/10 & 4/10 & 79/140 \\
      \textbf{ManiVID-3D} & 
      \textbf{18/20} & \textbf{15/20} & \textbf{11/20} & \textbf{16/20} & \textbf{18/20} & 
      \textbf{8/10} & \textbf{5/10} & \textbf{7/10} & \textbf{5/10} & \textbf{103/140}\\
      \bottomrule
    \end{tabular}
  }
\end{table*}

\subsection{Real-World Experiments}\label{sec real}

\begin{figure}[thbp] 
\raggedleft 
\includegraphics[width=0.48\textwidth]{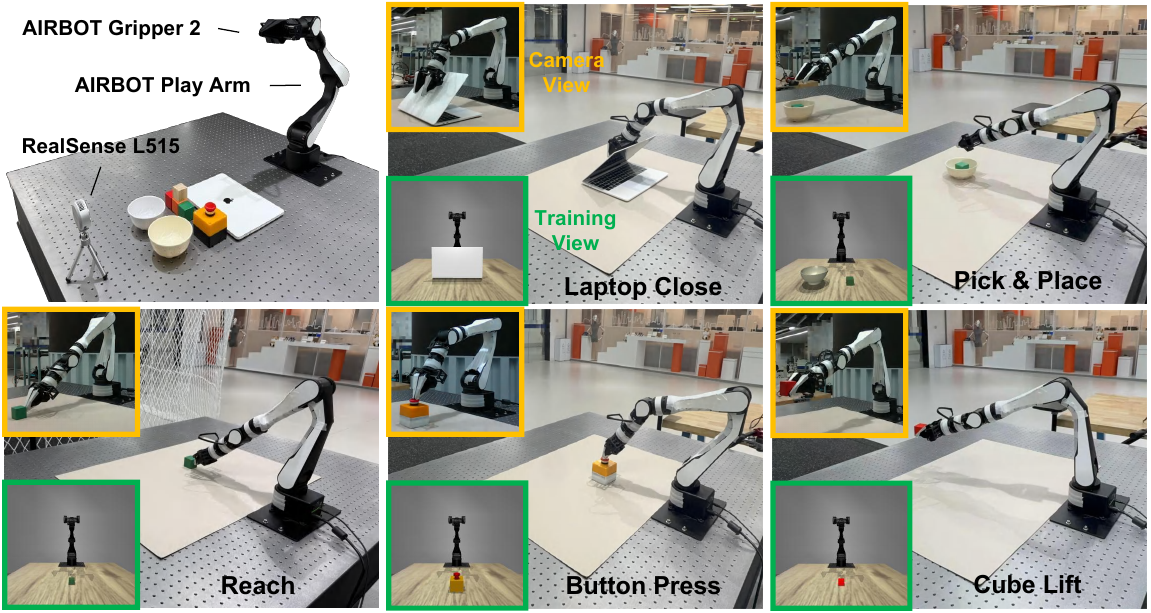}
\caption{\textbf{Real-world experiment setup and snapshots.} We conducted extensive real-world experiments using the AIRBOT Play arm with AIRBOT Gripper 2.}
\label{fig:6}
\end{figure}

As illustrated in Fig. \ref{fig:6}, we conduct real-world experiments by deploying our simulation-trained model on the AIRBOT Play arm + AIRBOT Gripper 2 platform in a zero-shot manner, with Maniwhere serving as the baseline. We select five tasks corresponding to the AIRBOT Play arm setup in simulation. For each task, we evaluate the algorithms from four distinct viewpoints with significant deviations from the reference perspective, performing five trials per viewpoint. Additionally, we test the generalization capabilities across four distinct scenarios: appearance variation, instance variation, lighting changes, and cluttered scenes (see the real-world snapshots in Fig. \ref{fig:6} for detailed setups). As demonstrated in Table \ref{tab:2real}, ManiVID-3D outperforms Maniwhere across all tasks with +17.1\% on average, showcasing its strong zero-shot sim-to-real transferability.

\begin{table}[!t]
  \centering
  \caption{\textbf{Ablation study on key components.} The results confirm the necessity of all designed components.}
  \label{tab:3 Ablation}
  \resizebox{0.48\textwidth}{!}{ 
      \begin{tabular}{l|ccc|c}
        \toprule
        Ablation & Cube Lift & Pick \& Place & Laptop Close & Average \\
        \midrule
        \textbf{ManiVID-3D} & \textbf{92.0$\pm$7.5} & \textbf{91.2$\pm$5.2} & \textbf{95.6$\pm$3.7} & \textbf{92.9} \\
        w/o ViewNet & 58.4$\pm$14.6 & 48.8$\pm$22.3 & 64.0$\pm$9.7 & 57.1 \\
        w/o $\mathcal{L}^{\text{inv}}$  & 87.2$\pm$8.4 & 86.8$\pm$7.2 & 91.6$\pm$5.2 & 87.9 \\
        w/o $\mathcal{L}^{\text{dep}}$\&$\mathcal{L}^{\text{orth}}$  ($\mathcal{\beta}$=0) & 89.8$\pm$6.5 & 86.2$\pm$8.5 & 92.6$\pm$4.6 & 89.5 \\
        w/o Curriculum ($\mathcal{\beta}$=1) &  90.8$\pm$7.4 &  89.6$\pm$6.5 &  95.2$\pm$4.8 &  91.9 \\
         w/o Curriculum ($\mathcal{\beta}$=10)  &  87.6$\pm$7.6 & 85.2$\pm$9.2 &  91.2$\pm$6.8 &  88.0 \\
        w/ 3D-STN \cite{3dstn} & 78.4$\pm$9.2 & 69.6$\pm$7.5 & 86.4$\pm$8.3 & 78.1 \\
        \bottomrule
      \end{tabular}
    }
\end{table}

\subsection{Ablation Study}\label{sec abla}
To validate the necessity of key designs in ManiVID-3D, we conduct ablation studies on its core components: ViewNet and the viewpoint-invariant disentanglement objective. Experiments on Cube Lift, Pick \& Place, and Laptop Close tasks evaluate four variants: (1) removing ViewNet, (2) eliminating the contrastive objective term $\mathcal{L}^{\text{inv}}$, (3) ablating disentanglement terms ($\mathcal{L}^{\text{dep}}$ and $\mathcal{L}^{\text{orth}}$), 
(4) removing the $\beta$ curriculum by fixing $\beta$ to a constant (we report $\beta=1$ and $10$; note that $\beta=0$ is identical to variant (3)),
and (5) replacing ViewNet with 3D-STN \cite{3dstn} integrated into RL training. 

Results in Table~\ref{tab:3 Ablation} show that ViewNet is critical—without viewpoint alignment, learning view-invariant representations becomes significantly harder, while 3D-STN trained in the self-supervised manner leads to incomplete alignment and performance degradation. Additionally, the contrastive objective ($\mathcal{L}^{\text{inv}}$, $\mathcal{L}^{\text{dep}}$, $\mathcal{L}^{\text{orth}}$) further enhances the model’s ability to handle view variations. Besides, replacing the $\beta$ curriculum with a fixed $\beta$ reduces overall success rates. Among constant settings, $\beta=1$ performs slightly better than $\beta=0$ and $10$, indicating that the disentanglement loss contributes positively to performance, but a curriculum is helpful to balance it with the view-invariant loss. Importantly, all constant-$\beta$ runs still converge to viable policies, indicating that training is not brittle to the choice of $\beta$. These findings confirm that each design in our method is essential for robust performance.

  
  
  

\begin{figure}[t!] 
\raggedleft 
\includegraphics[width=0.48\textwidth]{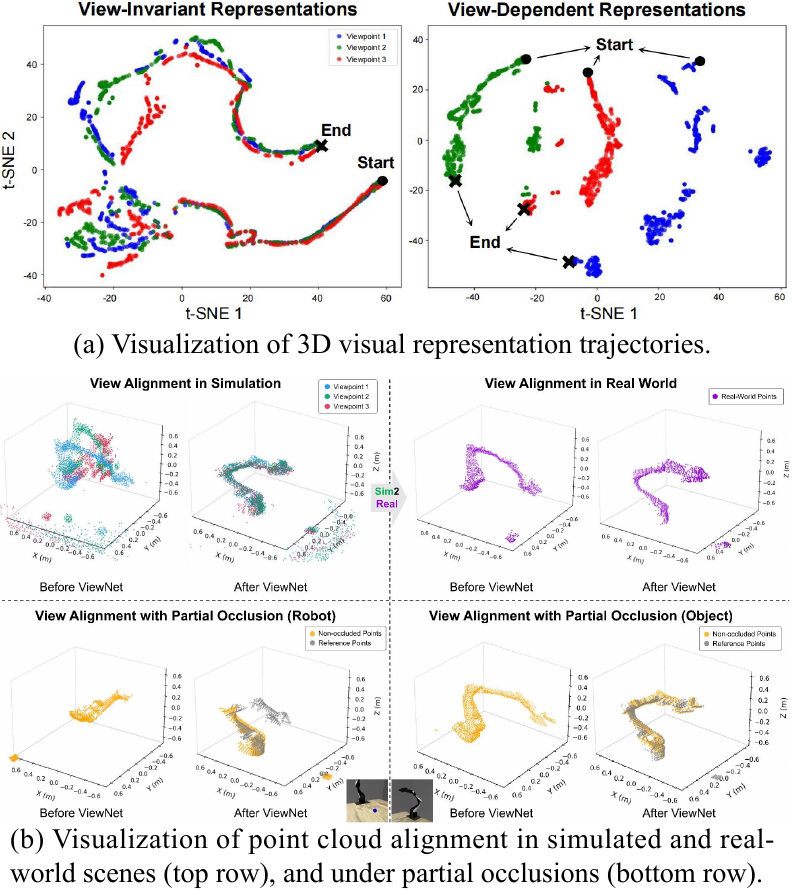}
\caption{\textbf{Qualitative analysis.} Point cloud distributions and representation trajectory consistency across viewpoints reveal our method's operating mechanism.}
\label{fig:7}
\end{figure}

\subsection{Qualitative Analysis}\label{sec QA}
In this section, we provide a qualitative analysis to explain ManiVID-3D’s strong performance in cross-viewpoint generalization. First, regarding viewpoint alignment, we visualize the raw point clouds from different viewpoints in both simulation and real world alongside their ViewNet-aligned counterparts as shown in Fig.~\ref{fig:7}(b). 
The results demonstrate that ViewNet achieves highly accurate viewpoint alignment while maintaining robust sim-to-real transferability, as evidenced by the clear-condition scenes (top row of  Fig.~\ref{fig:7}(b)). Moreover, by augmenting the training data with partially occluded point clouds, ViewNet shows strong robustness against diverse regional occlusions, as validated by the challenging cases in the bottom row of Fig.~\ref{fig:7}(b), including both robot occlusions and object occlusions.
Second, in terms of visual representation, we apply t-SNE visualization to the disentangled representations obtained from the same execution trajectory across different viewpoints. As shown in Fig.~\ref{fig:7}(a), ManiVID-3D successfully maps multi-view point cloud trajectories into two distinct spaces: (1) viewpoint-invariant representations, which cluster closely together, and (2) viewpoint-dependent representations, which remain well-separated. This confirms that our model effectively disentangles task-relevant features from viewpoint variations.

\section{Conclusion}

In this paper, we present ManiVID-3D, a novel generalizable view-invariant 3D visual RL framework with parallel efficient training for robotic manipulation. Our approach combines supervised view alignment with self-supervised feature disentanglement to achieve robust view invariance, while the introduced Efficient Batch Simulation with Rendering module significantly improves training efficiency. Our experiments demonstrate that ManiVID-3D outperforms existing baselines in both view robustness and sim-to-real transferability, while being more computationally efficient. 

\textbf{Future Work.} Future research directions include enhancing the framework's robustness against substantial noise clusters. We plan to explore the integration of adaptive denoising and RGB/tactile fusion to improve robustness.




\balance

\bibliographystyle{ieeetr}
\bibliography{ref2_0}

\ifCLASSOPTIONcaptionsoff
  \newpage
\fi

\end{document}